\documentclass[11pt,a4paper]{article}
\usepackage[hyperref]{naaclhlt2018}
\usepackage{times}
\usepackage{latexsym}

\usepackage{url}
\usepackage{tikz}
\usepackage{tabularx}
\usepackage{soul}
\usepackage{caption}
\usepackage{multirow}
\usepackage{pgfplots}
\aclfinalcopy 

\title{GWU NLP Lab at SemEval-2019 Task 3: \textit{EmoContext}: Effective Contextual Information in Models for Emotion Detection in Sentence-level in a Multigenre Corpus}

\author{Shabnam Tafreshi \\
  George Washington University \\
  Department of Computer Science \\
  Washington, DC \\
  {\tt shabnamt@gwu.edu} \\\And
  Mona Diab \\
  AWS AI\\
  George Washington University \\
  Department of Computer Science \\
  Washington, DC \\
  {\tt mtdiab@gwu.edu}}

\date{}

\begin{document}
\maketitle

\begin{abstract}
In this paper we present an emotion classifier model submitted to the SemEval-2019 Task 3: \textit{EmoContext}. The task objective is to classify emotion (\textit{i.e.} happy, sad, angry) in a 3-turn conversational data set. We formulate the task as a classification problem and introduce a Gated Recurrent Neural Network (GRU) model with attention layer, which is bootstrapped with contextual information and trained with a multigenre corpus. We utilize different word embeddings to empirically select the most suited one to represent our features. We train the model with a multigenre emotion corpus to leverage using all available training sets to bootstrap the results.  We achieved overall \%56.05 f1-score and placed 144.
\end{abstract}

\section{Introduction}
In recent studies, deep learning models have achieved top performances in emotion detection and classification. Access to large amount of data has contributed to these high results. Numerous efforts have been dedicated to build emotion classification models, and successful results have been reported. In this work, we combine several popular emotional data sets in different genres, plus the one given for this task to train the emotion model we developed. We introduce a multigenre training mechanism, our intuition to combine different genres are a) to augment more training data, b) to generalize detection of emotion. We utilize Portable textual information such as subjectivity, sentiment, and presence of emotion words, because emotional sentences are subjective and affectual states like sentiment are strong indicator for presence of emotion. \\
The rest of this paper is structured as followings: section \ref{mod} introduce our neural net model, in section \ref{exp} we explain the experimental setup and data that is been used for training and development sets, section \ref{res} discuss the results and analyze the errors, section \ref{rel} describe related works, section \ref{con} conclude our study and discuss future direction. 
\section {Model Description}
\label {mod}
\begin{figure}[!h]
\captionsetup{font=small}
\begin{center}
\includegraphics[scale=0.82]{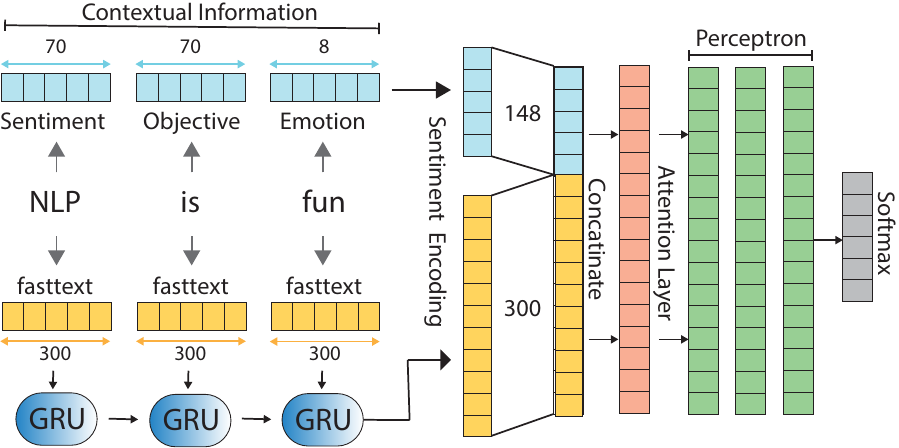}
\caption{GRU-Attention neural net architecture. In this model framework, context information are features generated from SentiWordNet and emotion lexicon. We use fasttext to show the embedding layer (we use ELMo too, but we do not show it in here). Features are presented to GRU and attention layer and the output of attention layer is sent to 3 perceptron layer. Last layer is a softmax layer to predict emotion labels. Model without contextual info, exclude the contextual info input, which we do not show in the architecture. }
\label{fig.2}
\end{center}
\end{figure}

Gates Recurrent Neural Network (GRU) \cite{cho2014learning,chung2015gated} and attention layer are used in sequential NLP problems and successful results are reported in different studies. Figure \ref{fig.2} shows the diagram of our model. \footnote{Data and system will be released upon the request.}\\\\
\textbf{GRU-} has been widely used in the literature to model sequential problems. RNN applies the same set of weights recursively as follow:
\begin{equation}
h_{t} = \textit{f}(W_{x_{t}} + Uh_{t-1} + b)
\end{equation}
GRU is very similar to LSTM with the following equations:  
\begin{equation} r_{t} = \sigma (W^r_{x_{t}} + U^rh_{t-1} + b^r) \end{equation}
\begin{equation} z_{t} = \sigma (W^z_{x_{t}} + U^zh_{t-1} + b^z) \end{equation}
\begin{equation} \hat{h_{t}} = \tanh (W_{x_{t}} + r_{t} \times U^{\hat{h}} h_{t-1} + b^{\hat{h}}) \end{equation}
\begin{equation} h_{t} = z_{t} \times h_{t-1} + (1 - z_{t}) \times \hat{h_{t}} \end{equation}
GRU has two gates, a reset gate \(r_{t}\), and an update gate \(z_{t}\). Intuitively, the reset gate determines how to combine the new input with the previous memory, and the update gate defines how much of the previous memory to keep around. We use Keras\footnote{\url{https://keras.io/}} GRNN implementation to setup our experiments. We note that GRU units are a concatenation of GRU layers in each task.\\\\
\textbf{Attention layer} - GRUs update their hidden state h(t) as they process a sequence and the final hidden state holds the summation
of all other history information. Attention layer \cite{bahdanau2014neural} modifies this process such that representation of each hidden state is an output in each GRU unit to analyze whether this is an important feature for prediction.\\\\
\textbf{Model Architecture -} our model has an embedding layer of 300 dimensions using \textit{fasttext} embedding, and 1024 dimensions using ELMo \cite{Peters:2018} embedding. \textit{GRU} layer has 70 hidden unites. We have 3 perceptron layers with size 300. Last layer is a \textit{softmax} layer to predict emotion tags. Textual information layers (explained in section \ref {modtext}) are concatenated with GRU layer as auxiliary layer. We utilize a dropout \cite{graves2013speech} layer after the first perceptron layer for regularization.
\subsection{Textual Information}
\label{modtext}
\textbf{Sentiment and objective Information (SOI)-} relativity of subjectivity and sentiment with emotion are well studied in the literature. To craft these features we use SentiwordNet \cite{baccianella2010sentiwordnet}, we create sentiment and subjective score per word in each sentences. SentiwordNet is the result of the automatic annotation of all the synsets of WORDNET according to the notions of \textit{positivity}, \textit{negativity}, and \textit{neutrality}. Each synset \textit{s} in WORDNET is associated to three numerical scores Pos(s), Neg(s), and Obj(s) which indicate how positive, negative, and objective (i.e., neutral) the terms contained in the synset are. Different senses of the same term may thus have different opinion-related properties. These scores are presented per sentence and their lengths are equal to the length of each sentence. In case that the score is not available, we used a fixed score 0.001.\\\\
\textbf{Emotion Lexicon feature (emo)-} presence of emotion words is the first flag for a sentence to be emotional. We use NRC Emotion Lexicon \cite{Mohammad13} with 8 emotion tags (\textit{e.i.} joy, trust, anticipation, surprise, anger, fear, sadness, disgust). We demonstrate the presence of emotion words as an 8 dimension feature, presenting all 8 emotion categories of the NRC lexicon. Each feature represent one emotion category, where 0.001 \footnote{empirically we observed that 0 is not a good initial value in neural net.} indicates of absent of the emotion and 1 indicates the presence of the emotion. The advantage of this feature is their portability in transferring emotion learning across genres.\\
\subsection{Word Embedding}
Using different word embedding or end to end models where word representation learned from local context create different results in emotion detection. We noted that pre-trained word embeddings need to be tuned with local context during our experiments or it causes the model to not converge. We experimented with different word embedding methods such as \textit{word2vec}, \textit{GloVe} \cite{pennington2014glove}, \textit{fasttext} \cite{mikolov2018advances}, and \textit{ELMo}. Among these methods \textit{fasttext} and \textit{ELMo} create better results.\\

\section{Experimental Setup}
\label{exp}
We split \textit{MULTI} dataset into 80\%,10\%,10\% for train, dev, and test, respectively. We use AIT and EmoContext (data for this task) split as it is given by SemEval 2018 and semEval 2019. We describe these data sets in details in the next section. All experiments are implemented using Keras \footnote{\url{https://keras.io/}} and Tensorflow \footnote{\url{https://www.tensorflow.org/}} in the back-end.
\subsection{Data}
\label{data}
We used three different emotion corpora in our experiments. Our corpora are as follows: a) A multigenre corpus created by \cite{tafreshi2018sentence} with following genres: \textit{emotional blog posts}, collected by \cite{aman2007identifying}, \textit{headlines} data set from SemEval 2007-task 14 \cite{strapparava2007semeval}, \textit{movie review} data set \cite {pang2005seeing} originally collected from Rotten tomatoes \footnote{https://www.rottentomatoes.com/} for sentiment analysis and it is among the benchmark sets for this task. We refer to this multigenre set as (MULTI), b) SemEval-2018 Affect in Tweets data set \cite{mohammad2018semeval} (AIT) with most popular emotion tags: \textit{anger}, \textit{fear}, \textit{joy}, and \textit{sadness}, c) the data set that is given for this task, which is 3-turn conversation data. From these data sets we only used the emotion tags \textit{happy}, \textit{sad}, and \textit{angry}. We used tag \textit{no-emotion} from MULTI data set as \textit{others} tag. Data statistics are shown in figures \ref{MULTIset}, \ref{AITset}, \ref{EmoContextset} .\\\\
\textbf{Data pre-processing - } we tokenize all the data. For tweets we replace all the URLs, image URLs, hashtags, @users with specific anchors. Based on the popularity of each emoticon per each emotion tag, we replace them with the corresponding emotion tag. We normalized all the repeated characters, finally caps words are replaced with lower case but marked as caps words.
\subsection{Training the Models} 
We have input size of 70 for sentence length, sentiment, and objective features and emotion lexicon feature has size 8. All these features are explained in section \ref {modtext} and are concatenated with GRU layer as auxiliary (input) layer. Attention comes next after GRU and have size 70. We select dropout of size 0.2. We select 30 epochs in each experiment, however, training is stopped earlier if 2 consecutive larger loss values are seen on evaluation of dev set. We use Adam \cite{kingma2014adam} optimizer with a learning rate 0.001. We use dropout with rates 0.2. The loss function is a categorical-cross-entropy function. We use a mini batch \cite{cotter2011better} of size 32. All hyper-parameter values are selected empirically. We run each experiment 5 times with random initialization and report the mean score over these 5 runs. In section \ref{res} we describe how we choose the hyper-parameters values.\\\\
\textbf{baseline-} in each sentence we tagged every emotional word using NRC emotion lexicon \cite{Mohammad13}, if any emotion has majority occurrence we pick that emotion tag as sentence emotion tag, when all emotion tags happen only once we randomly choose among them, when there is no emotional word we tag the sentence as \textit{others}. We only use the portion of the emotion lexicon that covers the tags in the task (\textit{i.e.} happy, sad, and angry).

\begin{table}
\captionsetup{font=small}
\begin{center}
\scalebox{0.7}{
\begin{tabular}{|l|l|l|l|l|l|l|}
      \hline
      Data set  & \#train & \#dev & \#test & total \\
      \hline
      MULTI  & 13776 & 1722 & 1722 & 17220 \\
      AIT & 6839 & 887 & 4072 & 11798 \\
      EmoContext  & 30160 & 2755 & 5510 & 38425 \\
      \hline
\end{tabular}}
\caption{Data statistics illustrating the distributions of the train, dev, and test sets across different data sets. } 
\label{emostat}
\end{center}
\end{table}

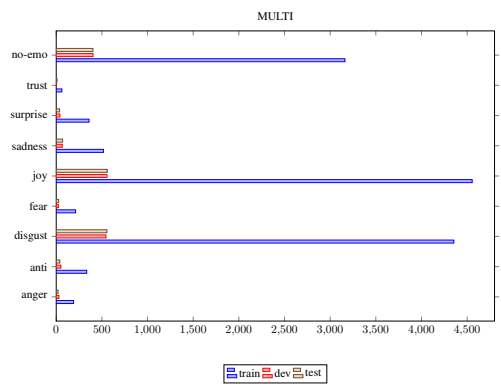
\begin{figure}[!h]
\captionsetup{font=small}
\begin{center}
\scalebox{0.4}{
\pgfplotstableread[row sep=\\,col sep=&]{
emotion  &train &dev &test\\
anger    &190   &30  &20\\
anti     &334   &51  &40\\
disgust  &4355	&545 &555\\
fear     &212	&24	 &27\\
joy      &4555	&556 &559\\
sadness  &517	&66  &70\\
surprise &359	&40  &37\\
trust    &62	&4	 &9\\
no-emo   &3162  &403 &402\\
    }\mydata
\begin{tikzpicture}
    \begin{axis}[
            xbar,
            bar width=.1cm,
            width=\textwidth,
            height=0.7\textwidth,
            legend style={at={(0.5,-0.15)},
            anchor=north,legend columns=-1},
            symbolic y coords={anger,anti,disgust,fear,joy,sadness,surprise,trust,no-emo},
            ytick=data,
            xmin=0,xmax=4800,
            title={MULTI},
        ]
        \addplot table[y=emotion,x=train]{\mydata};
        \addplot table[y=emotion,x=dev]{\mydata};
        \addplot table[y=emotion,x=test]{\mydata};
        \legend{train, dev, test}
    \end{axis}
\end{tikzpicture}}
\caption{MULTI data set - train, dev, test data statistic}
\label{MULTIset}
\end{center}
\end{figure}
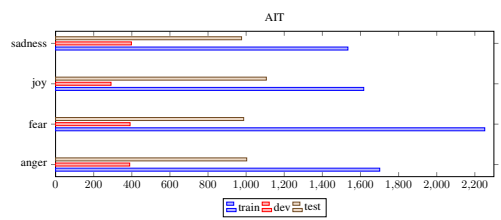
\begin{figure}[!h]
\captionsetup{font=small}
\begin{center}
\scalebox{0.4}{
\pgfplotstableread[row sep=\\,col sep=&]{
emotion	&train	&dev	&test\\
anger	&1701	&389	&1003\\
fear	&2252	&391	&987\\
joy		&1617	&291	&1106\\
sadness	&1534	&398	&976\\
    }\mydata
\begin{tikzpicture}
    \begin{axis}[
            xbar,
            bar width=.1cm,
            width=\textwidth,
            height=0.4\textwidth,
            legend style={at={(0.5,-0.15)},
            anchor=north,legend columns=-1},
            symbolic y coords={anger,fear,joy,sadness},
            ytick=data,
            xmin=0,xmax=2300,
            title={AIT},
        ]
        \addplot table[y=emotion,x=train]{\mydata};
        \addplot table[y=emotion,x=dev]{\mydata};
        \addplot table[y=emotion,x=test]{\mydata};
        \legend{train, dev, test}
    \end{axis}
\end{tikzpicture}}
\caption{AIT data set - train, dev, test data statistic}
\label{AITset}
\end{center}
\end{figure}
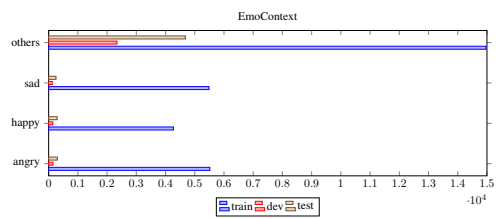
\begin{figure}[!h]
\captionsetup{font=small}
\begin{center}
\scalebox{0.4}{
\pgfplotstableread[row sep=\\,col sep=&]{
emotion	&train	&dev	&test\\
angry	&5516	&151	&299\\
happy	&4272	&144	&290\\
sad		&5487	&125	&252\\
others	&14965	&2338	&4678\\
    }\mydata
\begin{tikzpicture}
    \begin{axis}[
            xbar,
            bar width=.1cm,
            width=\textwidth,
            height=0.4\textwidth,
            legend style={at={(0.5,-0.15)},
            anchor=north,legend columns=-1},
            symbolic y coords={angry,happy,sad,others},
            ytick=data,
            xmin=0,xmax=15000,
            title={EmoContext},
        ]
        \addplot table[y=emotion,x=train]{\mydata};
        \addplot table[y=emotion,x=dev]{\mydata};
        \addplot table[y=emotion,x=test]{\mydata};
        \legend{train, dev, test}
    \end{axis}
\end{tikzpicture}}
\caption{EmoContext data set - train, dev, test data statistic}
\label{EmoContextset}
\end{center}
\end{figure}
\section{Results and Analysis}
\label{res}
The results indicates the impact of contextual information using different embeddings, which are different in feature representation. Results of class \textit{happy} without contextual features has \%44.16 by GRU-att-ELMo model, and \%49.38 by GRU-att-ELMo+F.\\
We achieved the best results combining ELMo with contextual information, and achieve \%85.54 f-score overall, including class \textit{others}. In this task we achieved \%56.04 f-score overall for emotion classes, which indicates our model needs to improve the identification of emotion. Table \ref{emoresultss} shows our model performance on each emotion tag. The results show a low performance of the model for emotion tag \textit{happy}, which is due to our data being out of domain. Most of the confusion and errors are happened among the emotion categories, which suggest further investigation and improvement. We achieved \%90.48, \%60.10, \%60.19, \%49.38 f-score for class \textit{others}, \textit{angry}, \textit{sad}, and \textit{happy} respectfully.\\
Processing ELMo and attention is computationally very expensive, among our models GRU-att-ELMo+F has the longest training time and GRU-att-fasttext has the fastest training time. Results are shown in table \ref{resultss} and table ref{emoresultss}
\begin {table}
\captionsetup {font=small}
\begin {center}
\scalebox{0.6}{
\begin {tabular}{|l|l|l|l|l|l|l|l|l|l|l|l|l|l|l|l|l}
\hline
Methods/Data set & \multicolumn{5}{c|}{\textbf{EmoContext}} \\ 
\hline
 & pr. & re. & f. & acc. & sp./\#epo.\\ 
\hline
Baseline & - & - & 46.20 & 46.20 & n.a. \\
GRU-att-fasttext & 88.12 & 81.24 & 83.44 & 80.84 & 103/14 \\
GRU-att-fasttext+F & 88.27& 84.47& 85.27& 82.07& 321/8 \\
GRU-att-ELMo  & 88.50 & 82.65 & 83.05 & 82.65 & 310/20 \\
GRU-att-ELMo+F & \textbf{88.61} & \textbf{84.34} & \textbf{85.54} & \textbf{83.62} & 960/28 \\
Context Results(emotion only) &54.28 & 57.93 & 56.04 & - & 960/28\\
\hline
\end {tabular}}
\caption {Results on the \textit{EmoContext} test sets. We report the mean score over 5 runs. Standard deviations in score are around 0.8. The experiments are demonstrating different embedding (\textit{i.e.} ELMo and fasttext), with features (F), which are \textit{emo} and \textit{SOI} explained in section \ref {modtext}}
\label {resultss}
\end {center}
\end {table}

\begin {table}
\captionsetup {font=small}
\begin {center}
\scalebox{0.7}{
\begin {tabular}{|l|l|l|l|l|l|l|l|l|l|l|l|l|l|l|l|l}
\hline
Emotion tags/Data set & \multicolumn{3}{c|}{\textbf{EmoContext}} \\ 
\hline
 & pr. & re. & f. \\ 
\hline
happy & 45.37 & 53.52 & 49.11\\
sad & 57.92 & 55.60 & 56.73\\
angry & 61.02 & 64.09 & 62.52\\
\hline
\end {tabular}}
\caption {Context results of each emotion tag.}
\label {emoresultss}
\end {center}
\end {table}
\section {Related Works}
\label{rel}
In semEval 2018 task-1, \textit{Affect in Tweets} \cite {mohammad2018semeval}, 6 team reported results on sub-task E-c (emotion classification), mainly using neural net architectures, features and resources, and emotion lexicons. Among these works \cite {baziotis2018ntua} proposed a Bi-LSTM architecture equipped with a multi-layer self attention mechanism, \cite {meisheri2018tcs} their model learned the representation of each tweet using mixture of different embedding. in \textit{WASSA 2017 Shared Task on Emotion Intensity} \cite {mohammad2017wassa}, among the proposed approaches, we can recognize teams who used different word embeddings: GloVe or word2vec \cite {he2017yzu,duppada2017seernet} and exploit a neural net architecture such as LSTM \cite {goel2017prayas,akhtar2017iitp}, LSTM-CNN combinations \cite {koper2017ims,zhang2017ynu} and bi-directional versions \cite {he2017yzu} to predict emotion intensity. Similar approach is developed by \cite{gupta2017sentiment} using sentiment and LSTM architecture. Proper word embedding for emotion task is key, choosing the most efficient distance between vectors is crucial, the following studies explore solution sparsity related properties possibly including uniqueness \cite{shen2018least, mousavi2017solution} . \\
\section{Conclusion and Future Direction}
\label{con}
We combined several data sets with different annotation scheme and different genres and train an emotional deep model to classify emotion. Our results indicate that semantic and syntactic contextual features are beneficial to complex and state-of-the-art deep models for emotion detection and classification. We show that our model is able to classify non-emotion (others) with high accuracy. \\
In future we want to improve our model to be able to distinguish between emotion classes in a more sufficient way. It is possible that hierarchical bi-directional GRU model can be beneficial, since these models compute history and future sequence while training the model. 

\bibliography{semeval2018}
\bibliographystyle{acl_natbib}

\end{document}